\documentclass{article}
\usepackage{amssymb}
\usepackage{graphicx}
\usepackage{float}
\usepackage{amsmath}
\usepackage{enumitem}
\usepackage[table]{xcolor}
\usepackage{placeins}
\usepackage{xcolor}
\usepackage[colorlinks=true, linkcolor=blue, urlcolor=blue]{hyperref}
\usepackage{subcaption}
\PassOptionsToPackage{numbers,sort}{natbib}
\usepackage[preprint]{corl_2025} % Uncomment for pre-prints (e.g., arxiv); This is like ``final'', but will remove the CORL footnote.

\usepackage{enumitem}
\setlist{nosep} % global: no extra vspace around itemize/enumerate

\usepackage[font=small,labelfont=bf]{caption}
\newcommand{\myparagraph}[1]{\vspace{0.01in}\noindent\textbf{#1}}

\definecolor{pastelGreen}{RGB}{226, 240, 217}
\definecolor{pastelRed}{RGB}{246, 221, 220}

\usepackage[compact]{titlesec}
\titlespacing*{\section}{0pt}{*0.8}{*0.65}
\titlespacing*{\subsection}{0pt}{*0.85}{*0.55}

\title{Reactive In-Air Clothing Manipulation with Confidence-Aware Dense Correspondence and Visuotactile Affordance}

\author{
  Neha Sunil$^{*1}$, Megha Tippur$^{*1}$, Arnau Saumell$^{2}$, Edward Adelson$^{1}$, Alberto Rodriguez$^{3}$\\
  $^{1}$Massachusetts Institute of Technology $^{2}$Prosper AI $^{3}$Boston Dynamics\\
  \texttt{<nsunil, albertor>@mit.edu <mhtippur, adelson>@csail.mit.edu}
  \vspace{-20pt}
\thanks{$^{}$Authors with equal contribution.}
\thanks{This work was done at MIT.}
}

\begin{document}
\maketitle

%===============================================================================
% NS: shorten so that teaser image can fit on this page
\begin{abstract}
    Manipulating clothing is challenging due to complex configurations, variable material dynamics, and frequent self-occlusion. Prior systems often flatten garments or assume visibility of key features. We present a dual-arm visuotactile framework that combines confidence-aware dense visual correspondence and tactile-supervised grasp affordance to operate directly on crumpled and suspended garments. The correspondence model is trained on a custom, high-fidelity simulated dataset using a distributional loss that captures cloth symmetries and generates correspondence confidence estimates. These estimates guide a reactive state machine that adapts folding strategies based on perceptual uncertainty. In parallel, a visuotactile grasp affordance network, self-supervised using high-resolution tactile feedback, determines which regions are physically graspable. The same tactile classifier is used during execution for real-time grasp validation. By deferring action in low-confidence states, the system handles highly occluded table-top and in-air configurations. We demonstrate our task-agnostic grasp selection module in folding and hanging tasks. Moreover, our dense descriptors provide a reusable intermediate representation for other planning modalities, such as extracting grasp targets from human video demonstrations, paving the way for more generalizable and scalable garment manipulation. See {\href{https://mhtippur.github.io/inairclothmanipulation/}{\textbf{\textcolor{cyan}{project website}}}} for demos. 
    
\end{abstract}
\vspace{-3mm}
% Two or three meaningful keywords should be added here
\vspace{-4mm}
%===============================================================================
% NS edit to version with extra label
\begin{figure}[H] 
    \centering
    \includegraphics[width=\linewidth]{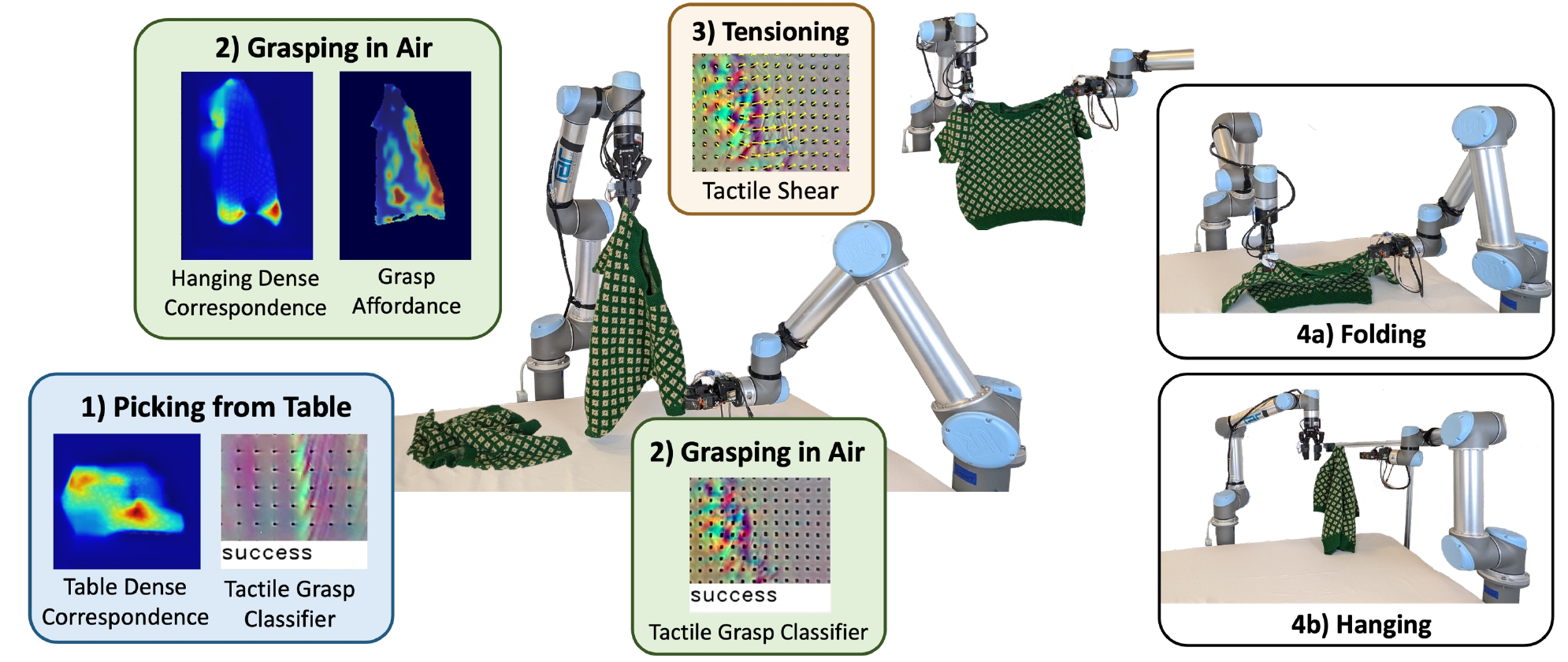}
    \vspace{-5mm}
    \caption{\textbf{Overview of visuotactile garment manipulation system.} Our framework integrates dense visual correspondence, visuotactile grasp affordance prediction, tactile grasp evaluation, and tactile tensioning for manipulating garments in highly-occluded configurations, both on a table-top and in-air. By leveraging a confidence-aware, reactive architecture and a task-agnostic representation, the system supports a variety of manipulation tasks, including folding and hanging.}
    \label{fig:teaser}
\end{figure}

\section{Introduction}
Deformable object manipulation is challenging because strategies developed for rigid objects rarely transfer. These objects have infinite-dimensional configuration spaces and uncertain dynamics. Garment manipulation adds further difficulty from intra-class variation, diverse material properties, and self-occlusion, especially in crumpled and suspended states where key visual cues may be hidden.

% Other rep techniques are insufficient
% NS: cite specific papers here
Existing cloth manipulation approaches often rely on flattening as an initial step \cite{Ganapathi2020, hoque2022learning, ha2022flingbot, clothfunnels}, expensive full-state estimation \cite{GarmentNets, chi2019cpd}, or task-specific grasp selection \cite{doumanoglou2014autonomous, aloha}. 
% Flattening-based methods reduce perceptual complexity but require additional preparatory actions. Full-state estimation is computationally expensive and brittle under severe occlusion. Task-specific grasp methods can achieve high success rates, but cannot effectively transfer knowledge to new tasks.
We propose a pose- and instance-agnostic, confidence-aware representation using dense visual descriptors to establish pixel-wise correspondences between deformed garments and canonical flat configurations. Trained on highly deformed states of detailed simulated shirts, our model operates directly on garments crumpled on a table or suspended in the air. Instead of the traditional contrastive loss used in prior garment dense correspondence work, we use a distributional loss that explicitly models garment symmetries and produces confidence estimates for each correspondence. We are able to handle configurations with heavier occlusion than those using contrastive dense descriptors because our system can defer low-confidence grasps until reliable visual information is available.

We integrate this correspondence representation into a visuotactile manipulation system, using high-resolution tactile sensing to (1) supervise grasp affordance learning, (2) validate grasp success during execution, and (3) enable closed-loop tensioning during folding. These components work together within a reactive framework that adapts folding and hanging strategies to garments of varying geometries, without requiring full-state estimation or flattening.

We make the following key technical contributions:
\vspace{-1mm}
\begin{itemize}[leftmargin=1.5em]
  \item \textbf{Parametrizable Simulated Dataset:} A custom dataset with realistic stitching features and parameterized variations to enable correspondences across different shirt geometries.
  \item \textbf{Dense Representation:} Pixel-wise correspondences across challenging garment states using a distributional loss to capture symmetries and provide confidence estimates.
  \item \textbf{Visuotactile Affordance:} A grasp affordance network trained in simulation and fine-tuned using tactile self-supervision to identify graspable regions with a single camera setup.
  \item \textbf{Garment Manipulation System:} A reactive visuotactile framework combining dense correspondences, affordances, and tactile sensing for confidence-aware in-air folding and hanging.
\end{itemize}
%===============================================================================
\section{Related Works}
Most previous cloth manipulation works focus on task-specific pipelines, including flattening \cite{ha2022flingbot, clothfunnels}, folding \cite{doumanoglou2014autonomous, maitin2010cloth}, dressing \cite{zhang2022learning, sun2024force}, and recently hanging \cite{aloha, chen2023learning, chen2024robohanger, chen2025}. These systems typically use incremental pick-and-place motions against a table \cite{Wu2019, Hoque2020, Ganapathi2020, lin2022learning}, and many focus on rectangular cloth, rather than garments.

Learning-based approaches can be quite successful at specific tasks. Labeling a real-world deformable object dataset is challenging \cite{Qian2020, chen2023learning}, so most learning works are trained in simulation. However, the sim2real gap remains a challenge. We address this for our grasp affordance network by extending \cite{Visuotactile}, fine-tuning using tactile classifiers to determine grasp success on the robot. Behavior cloning approaches \cite{aloha} have shown impressive results on tasks like tying shoelaces and hanging shirts, but require thousands of expert teleoperated demonstrations per task. In contrast, our system enables one- or few-shot generalization abilities and can reuse a shared object-centric representation across tasks.

\myparagraph{Perception and Representation} Early cloth manipulation work relies on corner detection or ridge detection \cite{yamazaki2011daily} to determine grasp points \cite{Willimon2011}. However, finding other more specific local features often requires first flattening the cloth \cite{Wu2019, hoque2022learning, lin2022learning, ha2022flingbot} or hanging it from specific grasp points \cite{doumanoglou2014autonomous, cusumano2011bringing, maitin2010cloth} to avoid self-occlusion. Some works determine the global state of the cloth \cite{tang2017cpd, chi2019cpd, GarmentNets}, but full-state inference is computationally expensive. In contrast, we use dense pixel-wise correspondences to directly localize task-relevant points in both deformed table-top and in-air configurations.

\myparagraph{Dense Descriptors} Dense visual descriptors have been used to learn pixel-level correspondences across object views~\cite{UniversalCorrespondenceNetwork, VisualDescriptorLearning}. \citet{DenseObjectNets} introduce dense object descriptors for task-agnostic manipulation, with follow-up work applying them to deformable objects~\cite{Ganapathi2020, RopeManipulationDOD, Unigarment}. Prior garment-specific applications use contrastive loss~\cite{Ganapathi2020, Unigarment}, but \citet{ganapathi2020mmgsd} use multimodal distributional loss \cite{petethesis} to model symmetry and uncertainty on ropes and square cloths. We extend this to garments, training on highly crumpled configurations and enabling in-air correspondence prediction, a capability not previously addressed, because of our ability to defer low-confidence actions. Our approach further differs from garment manipulation in \cite{Unigarment} because of our use of reactive control, made possible by confidence-aware descriptors and tactile feedback. We also discuss how our dense descriptors can act as an intermediate representation for different planning modalities, including learning from human video demonstrations. Additionally, ~\citet{huang2024rekep} uses DinoV2 \cite{dinov2} and a vision-language model to determine grasp points and constraints; our descriptors could be combined with a language model to find keypoint candidates to better support manipulation in more deformed states.
 
% \subsection{Tactile for Deformable Objects} \citet{Visuotactile} extends this work by sliding along the edge of the cloth. That paper also demonstrates the use of tactile sensing to supervise grasp success both for long-horizon task execution and training a grasp affordance network. We will use both of those abilities in this work.
%===============================================================================
%===============================================================================

\section{Methods}
\label{sec:methods}
\subsection{Dataset Generation in Simulation}

\begin{figure}[h!] 
    \centering
    \includegraphics[width=\linewidth]{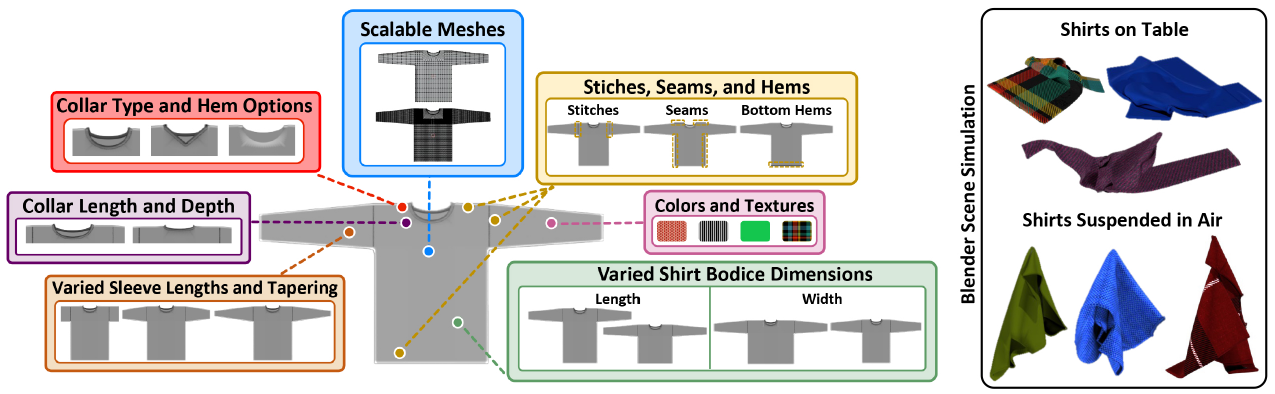}
    \vspace{-5mm}
    \caption{\textbf{Generating a simulated shirt dataset.} Blender 4.2 is used to simulate deformed shirts. Our animation pipeline allows flexibility in shirt geometries with the addition of realistic, key features like seams and hems often found on real shirts. A consistent vertex indexing across the shirt dataset is used, allowing alignment with a canonical template.}
    \label{fig:simulation}
\end{figure}

We use Blender 4.2 \cite{blender42} to simulate a wide variety of shirt geometries and deformations, generating a large RGB-D dataset (1500 scenes) for training. In addition to parameterizing the overall geometries and material properties, we use \cite{albisser2024sewingtoolbox} to incorporate hems, stitches, and sewing seams into our shirts to mimic realistic garments, enhancing visual realism and providing key features helpful for correspondence. Our method incorporates these finer details while preserving consistent vertex indexing across shirts, enabling descriptors to align with a canonical template regardless of geometry, without relying on sparse skeleton keypoints as in \cite{Unigarment}. \autoref{fig:simulation} shows some of the parameters and shirt configurations we randomize to generate our dataset.

Scene generation mimics real-world camera setups, with three cameras arranged radially around each shirt, with added pose noise and varied lighting conditions to enhance dataset diversity. For each suspended scene, a shirt is held from a random mesh point and the world coordinates and pixel locations of the deformed mesh vertices are saved. For each table scene, a randomly positioned flat shirt is repeatedly grasped from random points and repositioned multiple times. This setup captures rich, diverse data across garment shapes, configurations (suspended and table), and visual contexts, enabling robust correspondence learning between different poses and shirt instances. See \hyperref[subsec:7.1]{Appendix~\ref*{subsec:7.1}} for further simulation details.

\subsection{Dense Correspondence with Distributive Loss}

We aim to learn dense pixel-wise correspondences between images of garments in deformed and flattened configurations. Given an RGB image $I \in \mathbb{R}^{W \times H \times 3}$, we define a mapping $f : \mathbb{R}^{W \times H \times 3} \rightarrow \mathbb{R}^{W \times H \times d}$ that assigns a $d$-dimensional descriptor to each pixel in $I$. This descriptor space allows correspondences to be established by comparing descriptors across images.

\myparagraph{Contrastive Loss} Contrastive methods, as used by \cite{Ganapathi2020, DenseObjectNets, Unigarment}, supervise this mapping by sampling pairs of matching and non-matching pixels across images. For a query pixel $u_a = (x_a, y_a)$ in image $I_a$ and a candidate pixel $u_b = (x_b, y_b)$ in image $I_b$, the descriptor distance $D(I_a, u_a, I_b, u_b) = \| f(I_a)(u_a) - f(I_b)(u_b) \|_2$ is minimized for matching pairs and pushed beyond a fixed margin $M$ for non-matching pairs. This enforces one-to-one correspondences but struggles with ambiguities caused by symmetries or occlusions, which are common in deformable objects. Symmetric Pixel-wise Contrastive Loss (SPCL) \cite{ganapathi2020mmgsd} extends this approach to support symmetric correspondences, allowing multiple valid matches per query pixel. However, they found the results to be unstable, and the discrete matches resulted in discontinuity issues. We will compare our network to these contrastive baselines.

\myparagraph{Distributional Loss}
To address these limitations, we adopt the distributional formulation from \cite{ganapathi2020mmgsd}, which explicitly models uncertainty over correspondences. Rather than supervising individual descriptor pairs, the network predicts a full probability distribution over all possible matches for a query pixel. In contrast, the softmax over contrastive descriptor distances does not produce a true calibrated probability distribution. Formally, we define an estimator $\hat{p}_b(x_i, y_j | I_a, I_b, x_a, y_a)$ that outputs the probability that each pixel $(x_i, y_j) \in I_b$ corresponds to a given query pixel $(x_a, y_a) \in I_a$. This estimator is defined as:
\vspace{-2mm}
\begin{equation} \label{eq:estimated_probability_distribution}
    \begin{aligned}
        \hat{p}_b(x_i, y_j \mid I_a, I_b, x_a, y_a) &= 
        \frac{\exp\left(-\|f(I_a)[x_a, y_a] - f(I_b)[x_i, y_j]\|_2^2\right)}
        {\sum_{i', j'} \exp\left(-\|f(I_a)[x_a, y_a] - f(I_b)[x_{i'}, y_{j'}]\|_2^2\right)}
        &\forall (x_i, y_j) \in I_b
    \end{aligned}
\end{equation}
\vspace{-3mm}

The target distribution $q_b$ is a multimodal isotropic Gaussian defined over $I_b$, with standard deviation $\sigma$ and modes centered at the ground-truth correspondence pixels, allowing the network to represent multiple valid matches and capture ambiguities from symmetry.

\begin{figure}[H] 
    \centering
    \includegraphics[width=\linewidth]{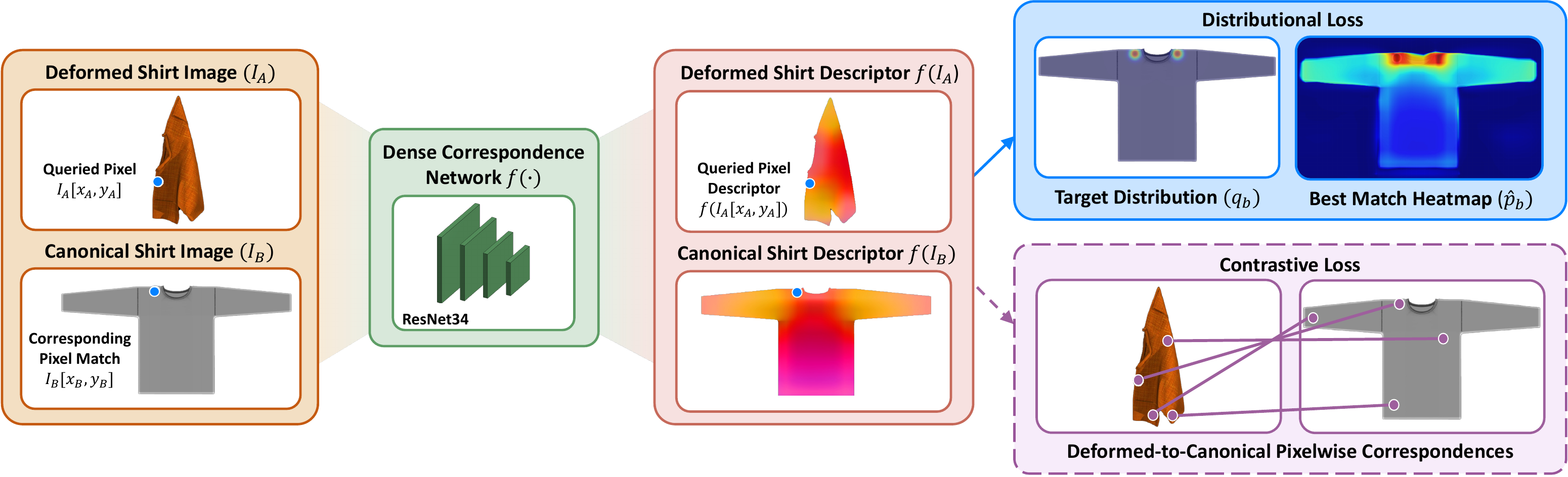}
    \vspace{-3mm}
    \caption{\textbf{Training dense correspondence in simulation.} Given two images $I_a$ and $I_b$, and a matching relation ($(x_a, y_a) \longleftrightarrow \{(x_b, y_b), (x_b', y_b')\}$), we train a CNN model $f$ to compute dense object descriptors. When supervising with distributional loss, we define a multimodal Gaussian target distribution $q_b$ with symmetrical modes over pixels corresponding to the queried point. We compute the probability distribution estimation $\hat{p}_{b_i}$ over image $I_b$ using $f(I_a)[x_a, y_a]$ and $f(I_b)$. Training minimizes the KL divergence between $q_b$ and $\hat{p}_{b_i}$. In the contrastive loss case, the model learns to push discrete pixel matches closer together in pixel space and non-matches further apart.}
    \vspace{-3.5mm}
    \label{fig:training}
\end{figure}

The descriptor mapping $f$ is implemented using ResNet34. The network is optimized by minimizing the Kullback-Leibler (KL) divergence between the predicted distribution $\hat{p}_{b_i}$ and the target distribution $q_{b_i}$ for each query pixel. Here, $\hat{p}_{b_i}$ is the predicted correspondence distribution over $I_b$ for the $i$-th query pixel (computed using \autoref{eq:estimated_probability_distribution}), and $q_{b_i}$ is the corresponding target distribution. \autoref{fig:training} shows a training example. 
At each iteration, we choose an image of a randomized deformed shirt and compare it to the canonical one. We query 50 randomly sampled points on the deformed shirt per iteration. 

Note that $I_b$ is always the canonical shirt image during training, meaning that we compute both the target and estimated distributions over the canonical shirt. A smooth Gaussian target distribution works over the canonical shirt because it does not have the occlusions and distortions of a deformed shirt. Defining the target distribution over the deformed shirt would be useful for training the network in both directions, but is unfeasible in this framework. 

\subsection{Visuotactile Grasp Affordance}
Training a general garment grasp affordance network is more challenging than for simpler deformable objects like towels. In \cite{Visuotactile}, the network was fine-tuned on a single towel with consistent material properties and dynamics. However, in this case, affordance must generalize across a wide range of geometries and material rigidities. As in \cite{Visuotactile}, we only use side grasps to reduce computational complexity. While grasp classifiers are trained for both grippers (as required by the larger system), affordance training is performed only for right-arm grasps, with left-arm affordance approximated by horizontally flipping inputs and outputs.

\myparagraph{Tactile Classifier}
To assess grasp quality, we train tactile classifiers to distinguish between successful grasps, grasps with too little fabric (which are empty or prone to slip), and grasps with a large excess of layers (more fabric than intended). We concatenate five evenly-spaced tactile depth images from the grasp attempt as input to our network. Our tactile datasets include 350 grasps across approximately 20 shirts, with limited augmentations, as implemented in \cite{Visuotactile} (two per input). 
% The classifier proved to be sensitive to translations and rotations, likely due to visible wrinkles on the gel surface through thin fabrics.

\myparagraph{Training Affordance in Simulation}
We use the same U-Net~\cite{ronneberger2015u} architecture as \cite{Visuotactile} for affordance prediction. The input to the network is a depth image of the hanging garment, and the output is an affordance heatmap over the image. Ground-truth affordance labels are computed per pixel via geometric analysis, leveraging full access to the cloth state in simulation. Specifically, each pixel is labeled based on gripper reachability, collision avoidance, and the number of fabric layers inside the gripper (restricted to two or fewer). These criteria are all explicitly checked in simulation, but the tactile classifier implicitly verifies these qualities on the robot. The simulated dataset consists of 300 unique cloth configurations, each rotated in increments of $30^{\circ}$, yielding a total of 3,600 images.

\myparagraph{Fine-tuning on the Robot}
We collect 8,500 grasp points for real-world fine-tuning to capture the greater variety of shirt dynamics and configurations compared to the simulated environment. Fine-tuning can easily overfit the real grasp dataset because the loss only applies to one pixel at a time. Furthermore, the tactile classifier cannot reliably determine whether the grasped region corresponds to the intended visual target. As a result, non-reachable pixels can yield positive tactile signals due to inadvertently grasping cloth in front of the target. To help address these challenges, our loss includes neighboring pixels to broaden supervision, along with regularization terms such as spatial smoothness penalties, simulation consistency constraints, and weight decay (see \hyperref[subsec:7.5]{Appendix~\ref*{subsec:7.5}}).

\subsection{In-Air Garment Manipulation}
\myparagraph{System Setup} Our bimanual system consists of two UR5 robots, both equipped with parallel-jaw grippers mounted with GelSight Wedge tactile sensors \cite{wang2021gelsight}. A Kinect Azure camera is used to capture RGB-D images. 

\myparagraph{Folding with Confidence-Based State Machine} Prior dense correspondence methods for garment folding rely on fixed pick points in canonical space \cite{Ganapathi2020, Unigarment}. While this simplifies planning, it fails when key features are occluded, distorted, or located in configurations unsuitable for grasping. 

Our system enables reactive in-air folding by dynamically selecting grasp points based on real-time confidence estimates and recovering from failures using tactile reactivity. The system starts by picking the shirt up from the table (looking for correspondence regions above a confidence threshold), and all subsequent grasps are performed in air.

% two modes: (1) starting with a random table grasp, followed by in-air pick point selection using a network trained solely on hanging garments, and (2) performing both the initial table grasp and subsequent in-air pick point selection using a single network trained on combined table and hanging data.

At each grasp attempt, the robot can query from three canonical regions (shoulder, sleeve, bottom) using our distributional dense correspondence network to generate confidence-weighted heatmaps. A grasp is executed only if both the correspondence confidence and grasp affordance (for in-air grasps) exceed predefined thresholds. Otherwise, the robot rotates the garment by $30^\circ$ and re-evaluates, ensuring robust grasp point selection across four folding strategies (shoulder-to-shoulder, bottom-to-bottom, sleeve-to-sleeve, sleeve-to-bottom). See \hyperref[subsec:7.2]{Appendix~\ref*{subsec:7.2}} for more details.

Grasp success is validated by tactile sensing (confirming fabric contact). If a grasp fails, the robot rotates and retries without releasing the garment. We use vision to ensure that the cloth is still in grip after moving the grippers. If no pixel meets the threshold requirements, the robot grasps the lowest available high affordance point to change configurations and encourage the cloth to unfurl. Once two confident grasp points are secured, the robot tensions the shirt (detecting shear via the average marker displacement on the tactile sensors) and can perform the rest of the folding motions open-loop, with the exception of using vision to align corners.

\myparagraph{Hanging} We demonstrate hanging by picking collar or shoulder points from the table and in the air. After securing both grasps, the robot moves open-loop to a peg. Hanging success is evaluated by grasp regions and whether the cloth stays on the peg.
	
%===============================================================================

\section{Results}
\label{sec:results}
\vspace{-4mm}
\begin{figure}[h!]
    \centering
    \includegraphics[width=1.0\textwidth]{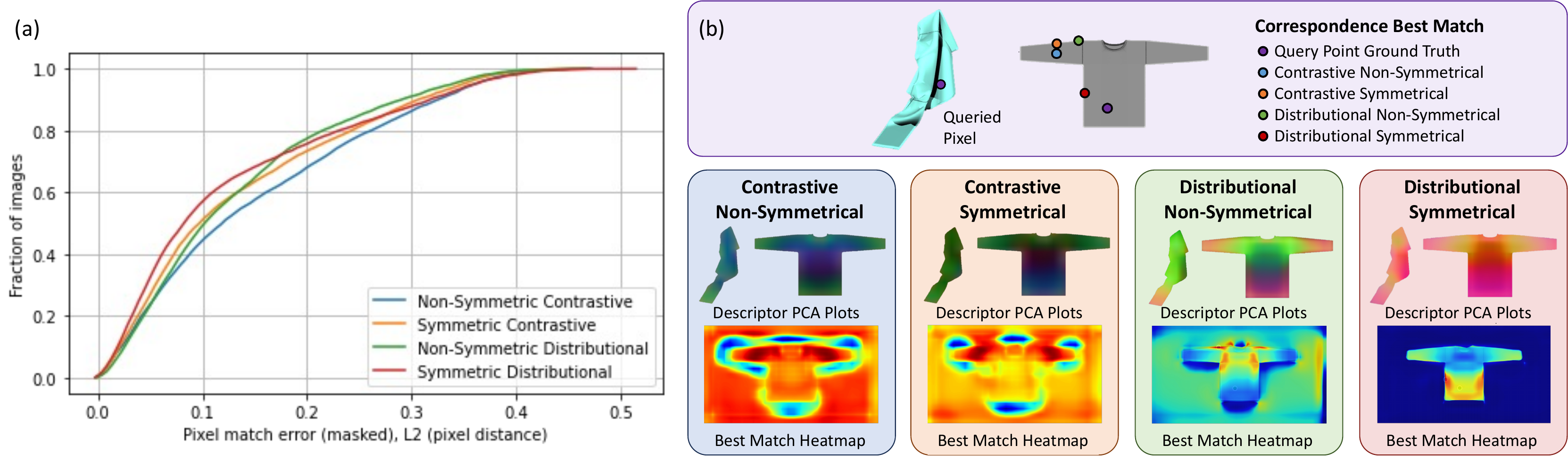}
    \vspace{-2mm}
    \caption{\textbf{Comparison of contrastive vs. distributional training, with and without symmetric supervision in simulation.} Plot (a) shows the cumulative fraction of image pixels whose predicted match is within a given pixel error threshold. Networks are trained on a combined dataset of suspended and table-top shirts, and evaluated on an unseen suspended test set; higher curves indicate better performance. Our symmetric distributional model performs the best at low error thresholds compared to baselines. For each network, (b) shows the predicted best pixel match for a queried point on a deformed simulated shirt, along with PCA visualizations of the dense descriptors in both the canonical and deformed states and the corresponding match heatmaps. Contrastive heatmaps are normalized between 0 and 1, while distributional heatmaps represent true calibrated probabilities.}
    \label{fig:contrastive}
\end{figure}

\myparagraph{Dense Correspondence} Most dense descriptor methods use contrastive one-to-one training \cite{Ganapathi2020, DenseObjectNets, Unigarment}, which fails to capture symmetries or spatial relationships beyond binary matches. Quantitative results (\autoref{fig:contrastive}) show similar cumulative pixel errors between contrastive and distributional models, but distributional models consistently outperform contrastive ones across nearly all error thresholds. Qualitatively, contrastive and non-symmetrical losses struggle with ambiguous structures, often confusing sleeves with the shirt bottom (as seen in PCA visualizations). In contrast, distributional loss supervises the model to predict a full probability distribution, enforcing spatial consistency. Explicit symmetry supervision further improves performance (\autoref{fig:contrastive}), especially at low error thresholds, by encouraging multimodal correspondences in symmetric regions.

We found that including random occlusions during training did not significantly affect performance in simulation, but helped improve performance on real data, likely due to masking artifacts. More detailed analysis of network parameters can be found in \hyperref[subsec:7.3]{Appendix~\ref*{subsec:7.3}}.

On real suspended shirt images, we evaluate our network by defining classification zones on the canonical shirt (see \hyperref[subsec:7.4]{Appendix~\ref*{subsec:7.4}}). When querying points from a suspended shirt (forward direction, as in training), the best suspended-only network classified the correct region 73.3\% of the time, while the best combined network (trained on both table and suspended data) achieved 62.2\% accuracy, while exhibiting lower overall confidence. Applying a confidence threshold, the combined network made correct, confidence-aware decisions (avoiding incorrect labels) 68.9\% of the time. In the inverse direction (querying from the canonical shirt), the combined network correctly identified the region 41.7\% of the time and made safe, confidence-aware decisions 70.8\% of the time. Some canonical points were occluded in the deformed image, making low confidence the correct outcome for these cases. On table scenes, the correct correspondence region was identified 70\% of the time, and a safe decision (either correct or low-confidence) was made 80\% of the time in 20 trials.

\myparagraph{Visuotactile Grasp Affordance}
Our tactile grasp classifier achieves 99.7\% accuracy on the right arm (used for tactile supervision) and 98.8\% on the left. Thin, flat shirts are the most challenging to classify. To evaluate affordance prediction, we collect 125 human-labeled grasp points where each point appeared potentially graspable to a human observer. We compare our fine-tuned affordance network against two baselines: (1) Sim2Real, trained in simulation and directly deployed, and (2) Real2Real, trained solely on robot data. Networks are evaluated offline using precision@k \cite{sanderson2010test}, a metric suitable for our unbalanced test set that avoids the need for a fixed threshold. We report precision@80, corresponding to the 80 successful grasps among the 125 test points. The results are 71.3\% for Sim2Real, 75.0\% for Real2Real, and 76.3\% for our fine-tuned network. Sim2Real performs worst due to discrepancies between simulated and real-world dynamics. While the fine-tuned and Real2Real networks achieve similar precision on the test set, qualitative analysis shows that Real2Real tends to be overconfident in incorrect predictions without the structure provided by the simulated network, particularly in less ambiguous cases not well-represented in the test set (see \hyperref[subsec:7.5]{Appendix~\ref*{subsec:7.5}}).

\vspace{1mm}
% RESULTS TABLE GOOD
\begin{table}[h!]

\centering
\renewcommand{\arraystretch}{1.2} % keep normal row spacing
\setlength{\tabcolsep}{4pt}
\begin{tabular}{|c!{\vrule width 1.75pt} c|c!{\vrule width 1.75pt} c|c !{\vrule width 1.75pt} c|c !{\vrule width 1.75pt}c|c|}
% \begin{tabular}{|c|c|c|c|c|c|c|c|c|}
\hline
\textbf{Category} & 
\multicolumn{2}{c!{\vrule width 1.75pt}}{\shortstack{\rule{0pt}{10pt}\textbf{Successful} \\ \textbf{Grasp (\%)}}} & 
\multicolumn{2}{c!{\vrule width 1.75pt}}{\shortstack{\rule{0pt}{10pt}\textbf{Corr.} \\ \textbf{Success (\%)}}} & 
\multicolumn{2}{c!{\vrule width 1.75pt}}{\shortstack{\rule{0pt}{10pt}\textbf{Low} \\ \textbf{Conf. (\%)}}} & 
\multicolumn{2}{c|}{\shortstack{\rule{0pt}{10pt}\textbf{Failed} \\ \textbf{Grasp (\%)}}} \\
\cline{2-9}
& Susp. & Comb. & Susp. & Comb. & Susp. & Comb. & Susp. & Comb. \\
\hline
Sleeve   & 60 & 40 & 80 & 60 & 10 & 10 & 30 & 50 \\
Bottom   & 40 & 10 & 90 & 90 & 40 & 80 & 20 & 10 \\
Shoulder & 40 & 60 & 100 & 100 & 60 & 20 & 0 & 20 \\
Collar   & 80 & 80 & 90 & 90 & 0 & 0 & 20 & 20 \\
\hline
\end{tabular}
\vspace{2mm}
\caption{\textbf{ (a) Grasping results using dense correspondence and grasp affordance across shirt categories for suspended and combined (suspended + table) dataset networks. } Low-confidence outcomes, where the shirt completes a full rotation without finding a grasp point, are not counted as successful or failed grasps. They are still included when calculating correspondence success, since both networks are trained to be confidence-aware. Failed grasps are categorized as either correspondence or affordance failures. Correspondence success rates exclude grasps that failed due to bad affordance predictions.}
\label{tab:grasping_results}
\end{table} 
% END RESULTS TABLE 

\vspace{-2mm}
\myparagraph{Combined System} We evaluate grasping performance across four garment regions (sleeve, bottom, shoulder, and collar) using two different correspondence networks: one trained solely on suspended shirts and another on a combined table and suspended dataset. For each category, we perform 10 grasp attempts per network, recording outcomes as success, failure, or below confidence threshold. Failures are further categorized as correspondence errors or affordance errors. In this experiment, we place the shirt in configurations where we expect graspable regions to emerge after rotation. Table~\ref{tab:grasping_results} summarizes rates for overall success, correspondence success (excluding bad affordance grasps), low-confidence rates, and total failure rates for each network and region.

The collar region consistently achieves higher confidence and success rates, likely due to its distinctive geometry. In contrast, the bottom region has the lowest confidence rates, reflecting its visual ambiguity and the increased difficulty of finding good affordance grasps from folding in on itself. The suspended network performs marginally better overall, but the combined network adds critical flexibility by supporting table grasps. Importantly, during folding, we query three candidate grasp points for the initial grasp, requiring confidence in only one to proceed. Subsequent grasps occur in easier, more unfurled configurations. 7 of 80 grasp attempts were affordance failures that the tactile classifier can recover from during task execution.

We found that our confidence-aware state machine was able to grasp viable folding points in 6 out of 10 trials. Of the 30 total grasps attempted during the course of our 10 folding trials, 6 were empty grasps successfully caught by the tactile classifier, immediately triggering recovery behaviors. Irrecoverable failure modes included correspondence failures, grabbing too much fabric, and grabbing diagonally across the shirt for sleeve-end grasps (despite masking out lowest points, see \hyperref[subsec:7.2]{Appendix~\ref*{subsec:7.2}}). Cloth slipping out was an occasional issue, but the system is able to recover. Without affordance fine-tuning, the folding success rate
dropped to 3 out of 10 trials, with an increase in cases of grabbing too much fabric caused by poor affordance, rather than poor correspondence. Our hanging system was successful in 7 out of 10 trials with all failures due to correspondence.

% CORRESPONDENCE FIGURE 
\begin{figure}[h!]
    \centering
    \includegraphics[width=1.0\textwidth]{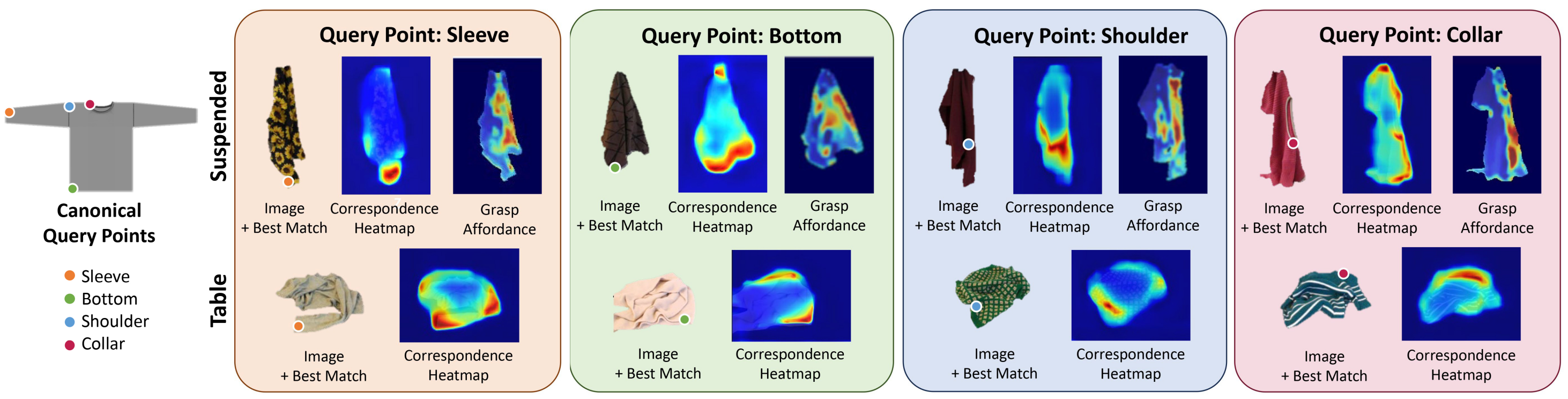}
    \vspace{0mm}
    \caption{\textbf{Correspondence and affordance heatmaps for real images.} We show examples for both suspended and table configurations, with correspondence probability maps for four query types: sleeve, shoulder, collar, and bottom. For suspended images, we also show the grasp affordance heatmap. In the robot system, grasp points are selected where both correspondence and affordance exceed predefined confidence thresholds. Note that while training queries points on the crumpled shirt, the robot queries points on the canonical image.}
    \label{fig:qual_results}
\end{figure}
%===============================================================================

%===============================================================================

\begin{figure}[b!]
    \centering
    \includegraphics[width=\textwidth]{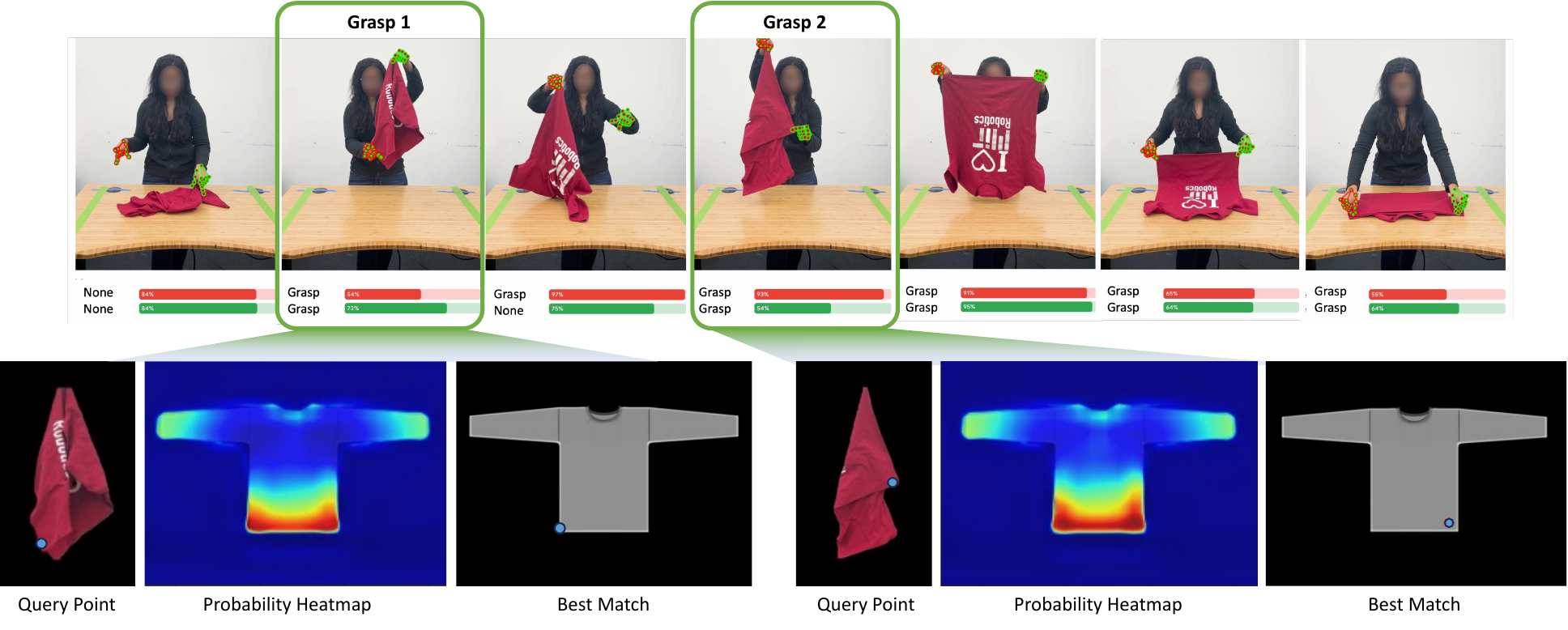}
    \vspace{-1mm}
    \caption{\textbf{Extracting grasp points from human video demonstrations.} We track hand gestures throughout the video to identify key moments. For each key frame, we use the tracked hand position to define a query point and retrieve the corresponding location on the canonical shirt using our dense correspondence model. This approach enables folding demonstrations to be interpreted as robot-executable instructions via our dense visual representation.}
    \label{fig:human_demo}
\end{figure}

\section{Conclusion}
\label{sec:conclusion}

We present a reactive visuotactile system for garment manipulation that integrates dense visual correspondence, visuotactile grasp affordance, confidence-aware planning, and tactile feedback. Unlike prior work constrained to table-top picking or reliant on flattening, our system supports in-air garment manipulation directly from crumpled states, guided by dense correspondences—a capability not previously demonstrated in the field. This enables more flexible, human-like manipulation.

A core insight of our work is the importance of confidence-driven reactivity: by deferring low-confidence actions and using tactile sensing for validation and correction, the system maintains robustness under severe occlusion and uncertainty. This closed-loop approach bridges the gap between global visual context and local contact feedback, enabling reliable control even when full object geometry is not observable.

Beyond task execution, our dense, confidence-aware representation serves as a generalizable intermediate layer for higher-level planning frameworks. It provides a foundation for extracting grasp targets from human video demonstrations (See \autoref{fig:human_demo} and \hyperref[subsec:7.6]{Appendix~\ref*{subsec:7.6}}), and has the potential to interface with vision-language models~\cite{huang2024rekep} or symbolic planners. These directions open the door to scalable, semantically-informed manipulation systems capable of adapting across garments, tasks, and contexts.

%===============================================================================

\section{Limitations}
\label{sec:limitations}

While our system demonstrates strong potential for in-air garment manipulation, several areas present opportunities for further development. First, the generalizability of the dense correspondence network is limited by the features available in simulation. Although we incorporated realistic details such as seams, hems, and varied necklines, other common garment features—like hoods, buttons, zippers, and mixed patterns—are not yet modeled. Some of these could be added in future dataset expansions, while others may require advances in simulation tools.  On out-of-distribution shirts (see \hyperref[subsec:7.4]{Appendix~\ref*{subsec:7.4}}), the network still captures general structure, but with lower confidence. New garment classes (e.g., trousers) would require new simulations, which would not require a complete overhaul of the simulation methodology, but would need to be updated to account for the different geometries, hems, and zippers. However, even in these cases, the overall grasp selection pipeline remains relatively unchanged, but would likely require a new confidence threshold or additional fine-tuning.  

Second, we are able to achieve this performance with a single camera and exclusively side approach grasps, but expanding to additional viewpoints and enabling more grasp approach angles could improve coverage to access more high correspondence regions. Incorporating temporal information could further enable the system to track keypoints as they become accessible, supporting more flexible planning.

Finally, although the system is confidence-aware, the network occasionally overestimates its certainty in challenging configurations. We experimented with auxiliary confidence prediction and KL-divergence metrics, but these did not significantly improve failure detection. Improving uncertainty estimation remains an important direction for future work.
%===============================================================================

\clearpage
\acknowledgments{
We would like to thank Sangbae Kim for helping provide the computing resources used in this work. 

This material is based upon work supported by the National Science Foundation Graduate Research Fellowship Program (NSF GRFP), the Toyota Research Institute (TRI), and the Amazon Science Hub. 
}

%===============================================================================

% no \bibliographystyle is required, since the corl style is automatically used.
\bibliography{references}  % .bib

%===============================================================================
\newpage
\section{Appendix}
\label{sec:appendix}

\subsection{Blender Simulation Parameters} 
\label{subsec:7.1}
We provide additional details on the Blender scene setup and parameters used to generate our combined shirt dataset (including both suspended in-air and on-table configurations). The ratios of shirt features are selected to loosely reflect the distribution of shirts we test on the real system. Rendering 50 scenes with these parameters takes 10 hours on an NVIDIA RTX 4090 GPU. 

\begin{table}[h]
\centering
\renewcommand{\arraystretch}{1.3}
\begin{tabular}{|l|p{9cm}|}
\hline
\multicolumn{2}{|c|}{\cellcolor{gray!30}\textbf{Blender 4.2 Simulated Shirt Scene Dataset Parameters}} \\
\hline
\multicolumn{2}{|c|}{\cellcolor{yellow!20}\textbf{Scene Parameters}} \\
\hline
Shirt Suspended in Air Scenes & 1000 scenes \\
\hline
Shirt on Table Scenes & 500 scenes \\
\hline
Cameras Rendered per Scene & 3 cameras \\
\hline
Fabric Quality Steps & 10 \\
\hline
Render Quality & 64 \\
\hline
\multicolumn{2}{|c|}{\cellcolor{blue!20}\textbf{Shirt Parameters}} \\
\hline
Mesh Vertex Density & 2922 \\
\hline
Shirt Thickness & 0.4 mm \\
\hline
Sleeve Length Ratio in Dataset & 65\% short sleeve, 35\% long sleeve \\
\hline
Neck Type Ratio in Dataset & 80\% U-Neck, 20\%V-Neck \\
\hline
Collar Hem Ratio in Dataset & 80\% collar hems, 20\% without collar hems \\
\hline
Bottom Hem Ratio in Dataset & 70\% without bottom bodice hems, 30\% bottom bodice hems \\
\hline
Shirt Stiffness Range & Uniformly sampled between [7, 15] \\
\hline
Shirt Damping Range & Uniformly sampled between [5, 7] \\
\hline
\end{tabular}
\vspace{0.5em}
\caption{Scene parameters used for dataset generation in Blender 4.2.}
\label{tab:shirt_scene_params}
\end{table}

\subsection{Folding with Confidence-Based State Machine}
\label{subsec:7.2}
\begin{figure}[h]
    \centering
    \includegraphics[width=1.0\textwidth]{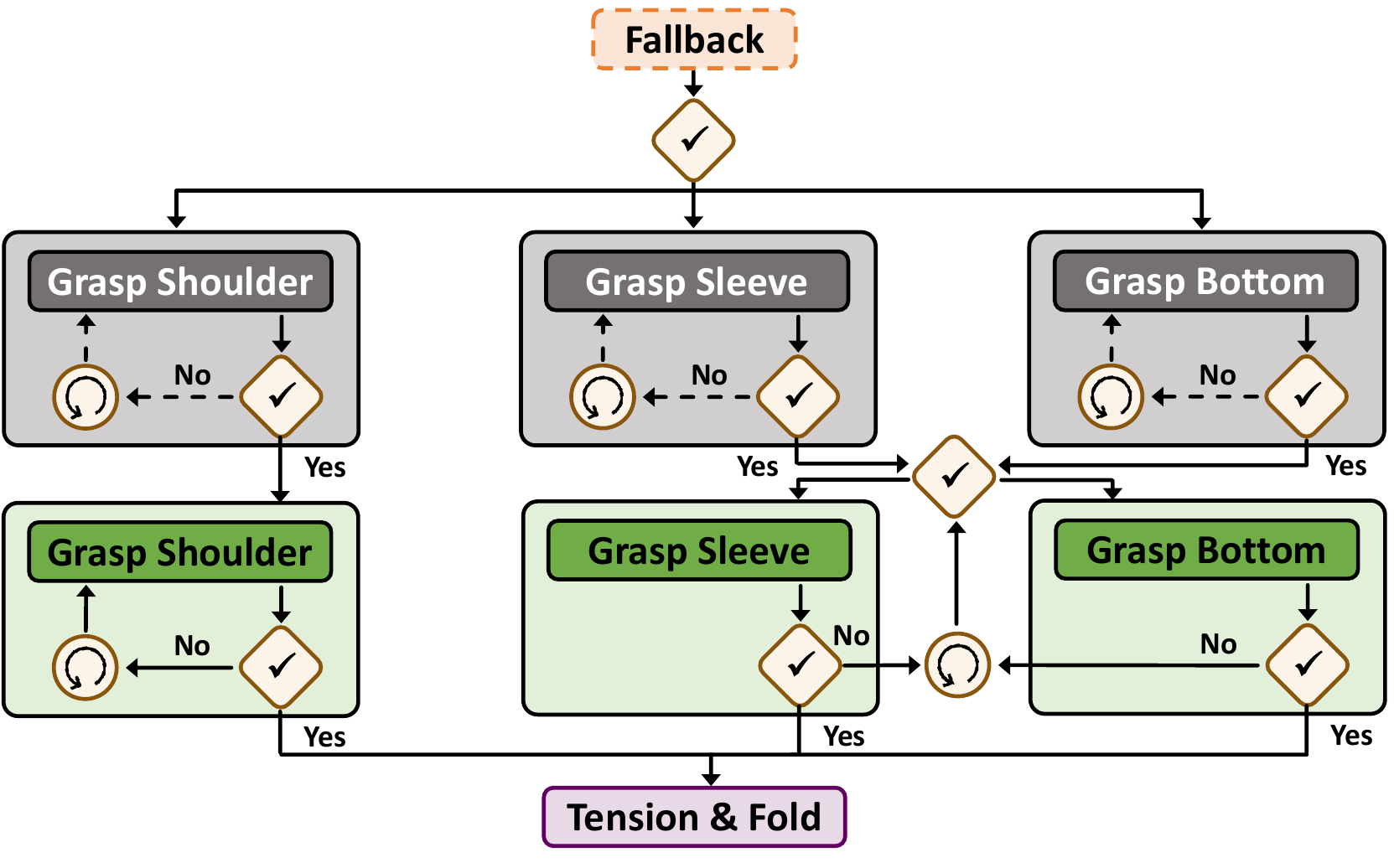}
    \caption{\textbf{Confidence-based state machine for folding strategy.} The robot dynamically chooses between folding strategies based on which points are visible and graspable. The initial grasp occurs on table, where the fallback strategy for low confidence is grasping the highest point. All subsequent grasps are attempted in air. The robot only attempts a grasp if correspondence confidence and grasp affordance exceed predefined thresholds. If no point is graspable, the robot rotates. If the robot completes a full rotation, the new fallback option is grabbing the lowest graspable point to help unfurl the cloth. Once two successful grasps are made, the robot tensions the cloth and folds.}
    \label{fig:state_machine_fold}
    \vspace{5mm}
\end{figure}

We allow the robot to choose the most appropriate folding pick points based on which points it can confidently identify and grasp. \autoref{fig:state_machine_fold} shows the four different folding strategies (shoulder to shoulder, bottom to bottom, sleeve to sleeve, sleeve to bottom). Bottom refers to the bottom corner of the shirt, and sleeve refers to the bottom edge of the sleeve. The system starts by picking the shirt up from the table (looking for high-confidence correspondence regions), and all subsequent grasps are performed in air. 

At each grasp attempt, the robot can query from three canonical regions (shoulder, sleeve, bottom) using our distributional dense correspondence network to generate confidence-weighted heatmaps. A grasp is executed only if both the correspondence confidence and grasp affordance (for hanging grasps) exceed predefined thresholds. Grasp success is validated by our tactile classifier (confirming fabric contact). If no grasp is attempted or the grasp attempt fails, the robot rotates the garment by $30^\circ$ and re-evaluates. In cases where symmetry matters (e.g. grabbing the sleeve and end on same side of the shirt), we use the heuristic that the opposite corner features would be the lowest point, and therefore we mask out the bottom. If no pixel meets the threshold requirements, the robot grasps the lowest available high affordance point to change configurations and encourage the cloth to unfurl.

The very first grasp attempt is done on the table. If no high correspondence point is found within the robot's workspace, the robot's fallback strategy is to grasp the highest point. All subsequent grasps are performed in air. The robot continues switching arms until it has two successful grasps.

Once the shirt is grasped by two keypoints, the robot pulls the shirt until it is tensioned. We use shear as measured by marker tracking on the tactile sensor as an indication for when the shirt is in tension. Then, the robot brings the lifted shirt to one end of the workspace, lowers it to the table, lowers the grippers to the other end of the table while resting half the shirt, then folds the shirt over as the grippers return to the first side of the workspace. The robot uses vision to align the corners in the final folding motion.

\begin{figure}[h!]
    \centering
    \includegraphics[width=\textwidth]{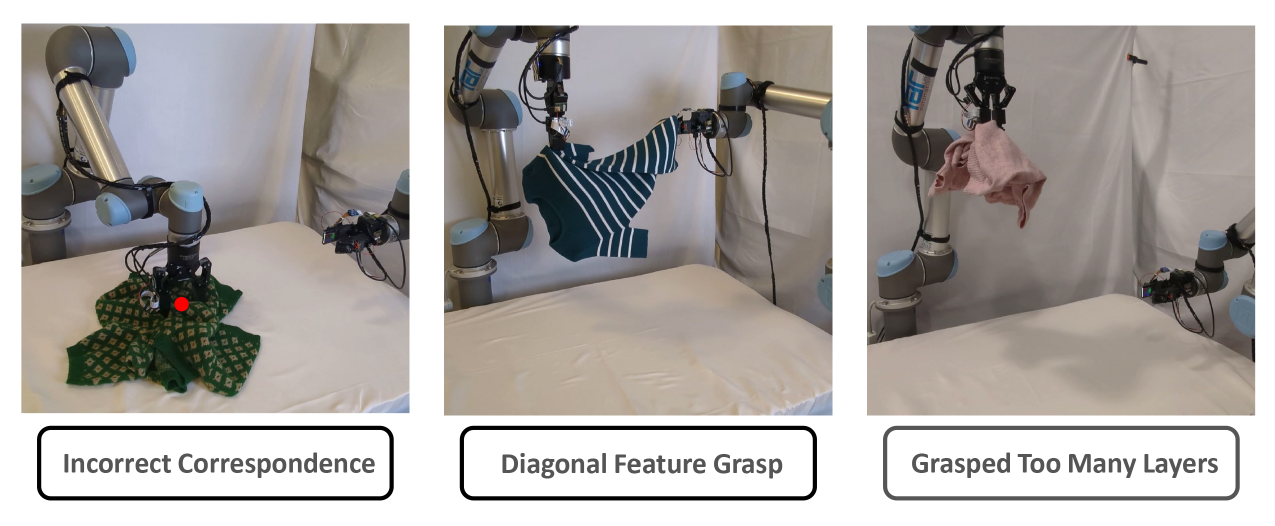}
    \caption{\textbf{Irrecoverable Failure Modes of Folding.} Though the confidence-based state machine is able to recover from mistakes in folding, some cases are unaccounted for and irrecoverable in the system. Incorrect correspondence grasps, picking the correct feature but on the wrong side, and grasping too much cloth are some of the failure cases. }
    \label{fig:bad_folds}
\end{figure}

Even with the confidence-based state machine, however, irrecoverable failure modes still occur. \autoref{fig:bad_folds} shows examples of these cases. Correspondence failures that result in grasps of internal points on the shirt (such as the body), grasping the correct feature but on the opposite side of the shirt, and grasping too many layers of fabric are some examples of failures that occur while folding. 

% MHT - Take Out this Table If We Decide to Leave The Other One In the Main File
We break down the failure cases by task:
\vspace{2mm}

\begin{table}[h]
\centering
% \scriptsize
\setlength{\tabcolsep}{5pt}
\renewcommand{\arraystretch}{1.1}
\begin{tabular}{|l|c|c|}
\hline
\textbf{Failure Breakdowns of 10 Trials} & \textbf{Folding} & \textbf{Hanging} \\
\hline
Incorrect Correspondence & 1 & 3 \\
Diagonal Feature Grasp   & 2 & 0 \\
Grasped Excess Layers    & 1 & 0 \\
\hline
\rowcolor{pastelRed}
\textbf{Total Failures}  & \textbf{4/10} & \textbf{3/10} \\
\hline
\rowcolor{pastelGreen}
\textbf{Total Successes}  & \textbf{6/10} & \textbf{7/10} \\
\hline
\end{tabular}
\label{tab:fail_modes}
\end{table}
\vspace{1mm}

Recoverable failures include affordance failures leading to insufficient cloth in the grip and the cloth slipping out of the grip. Our tactile classifier informs the system if each grasp is successful. We use vision to ensure that the cloth is still in grip after moving the grippers.

\newpage
\subsection{Dense Correspondence Network Parameters}
\label{subsec:7.3}
The mapping function $f$ that generates the dense descriptor space is implemented as a 34-layer ResNet (pretrained on ImageNet) with a stride of 8 for computational efficiency (as in \cite{DenseObjectNets}). Bilinear upsampling is applied to the network’s feature maps to align the output descriptor maps with the input image size (540×960 pixels). We train each of our final networks for approximately 10,000 iterations, which takes under 2 hours on an NVIDIA RTX 4090 GPU.

\begin{figure}[h!]
    \centering
    \includegraphics[width=0.7\textwidth]{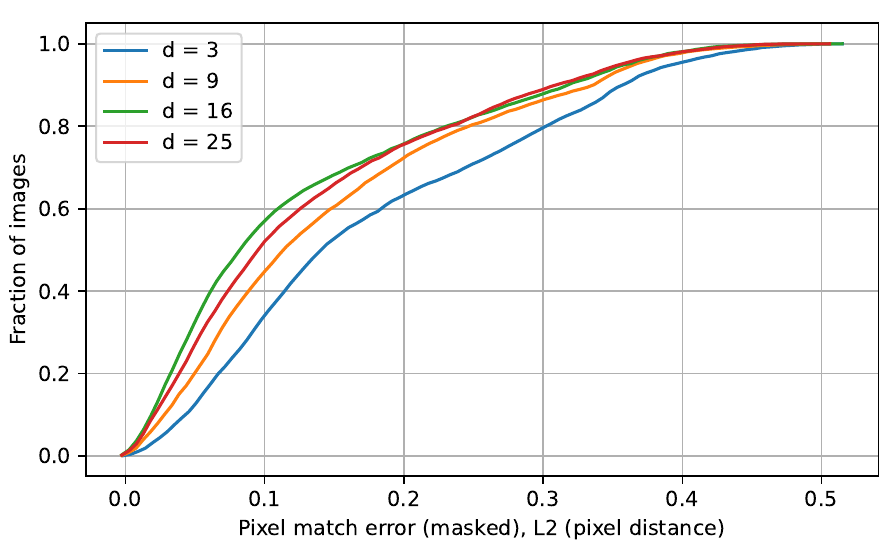}
    \caption{\textbf{Cumulative pixel match error across different descriptor dimensions ($d$) evaluated on the simulated test set.} The network was trained on a combined dataset of hanging and table shirts. A descriptor size of $d=16$ provides the best trade-off between representational capacity and generalization, outperforming both smaller ($d=3$, $d=9$) and larger ($d=25$) dimensions.}
    \label{fig:descriptor_size}
\end{figure}

\myparagraph{Hyperparameter Tuning} We conducted a series of hyperparameter experiments to optimize the performance of our dense correspondence network. A key parameter was the descriptor dimension $d$, which controls the capacity of the embedding space. As shown in \autoref{fig:descriptor_size}, we tested dimensions of 3, 9, 16, and 25. A descriptor size of $d=16$ consistently outperformed smaller and larger alternatives, striking a balance between sufficient representational capacity and generalization. Lower dimensions (e.g., $d=3$) lacked expressivity, while higher dimensions (e.g., $d=25$) did not offer noticeable improvements and introduced potential overfitting. Additionally, larger networks require more computation time.

\begin{figure}[h!]
    \centering
    \includegraphics[width=0.7\textwidth]{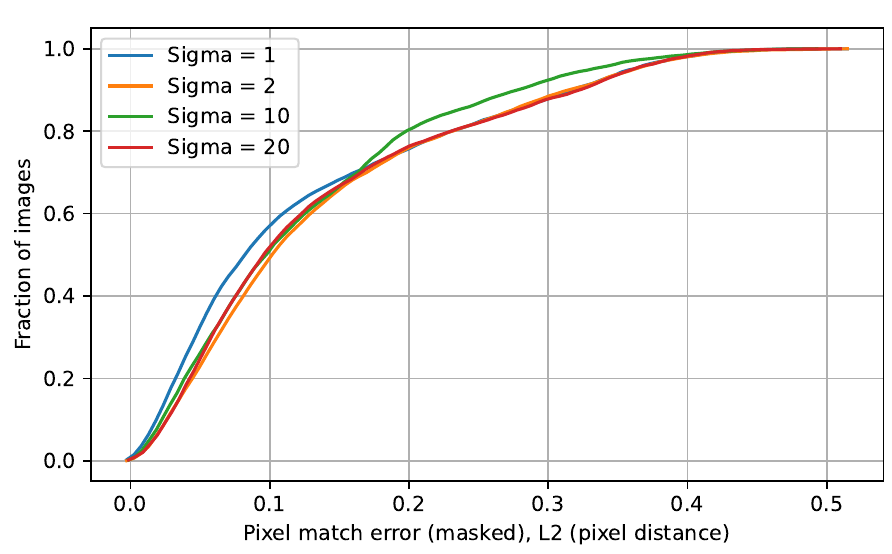}
    \caption{\textbf{Cumulative pixel match error for different Gaussian $\sigma$ values used in the distributional loss target.} The network was trained on a combined dataset of hanging and table shirts. Smaller $\sigma$ values (e.g., $\sigma=1$) produce sharper distributions and yield slightly better accuracy in simulation, but larger $\sigma$ values improve generalization to real-world data by promoting smoother gradients in the descriptor space.}
    \label{fig:sigma}
\end{figure}

We also evaluated the effect of $\sigma$, the standard deviation of the Gaussian used for the distributional loss target. \autoref{fig:sigma} shows performance across $\sigma$ values of 1, 2, 10, and 20. While $\sigma = 1$ yielded sharper distributions and slightly better accuracy in simulation, we found that larger $\sigma$ networks generalized better to real-world data. We hypothesize that broader Gaussians produce smoother gradients across the descriptor space, which in turn leads to more stable and consistent correspondence predictions. This smoothing effect could help mitigate sensitivity to local noise, masking artifacts, or out-of-distribution lighting. Sharper distributions (from smaller $\sigma$) can lead the network to overfit to high-frequency details in the simulated data, which don’t transfer well to real-world images.

\myparagraph{Model and Dataset Design Choices}
During early testing, we also experimented with several architectural variations. We evaluated higher-resolution ResNets and a DINOv2 backbone for the mapping function $f$, but found that DINOv2 performed significantly worse given our limited dataset size, and the higher-resolution ResNets did not yield noticeable improvements in correspondence accuracy. To improve confidence estimation, we attempted to train a separate confidence head; however, this approach did not reliably predict correspondence accuracy.

Additionally, our initial training dataset lacked hem and seam details, which led to poor differentiation between sleeve and torso ends when applied to real garments. Including these structural details in later dataset versions improved real-world performance. For hanging datasets with 1000 scenes, our best descriptor network with seams had a 73.3\% classification success rate in the forward direction. Without seams, the best network had a 42.2\% success rate. This network often misclassified sleeve regions as bottom regions.

We also experimented with incorporating depth information alongside RGB inputs but observed no significant gains. This suggests that in our cloth manipulation tasks, texture and color cues dominate the correspondence signal, and depth alone does not meaningfully contribute to distinguishing garment regions.

\begin{figure}[h!]
    \centering
    \includegraphics[width=0.7\textwidth]{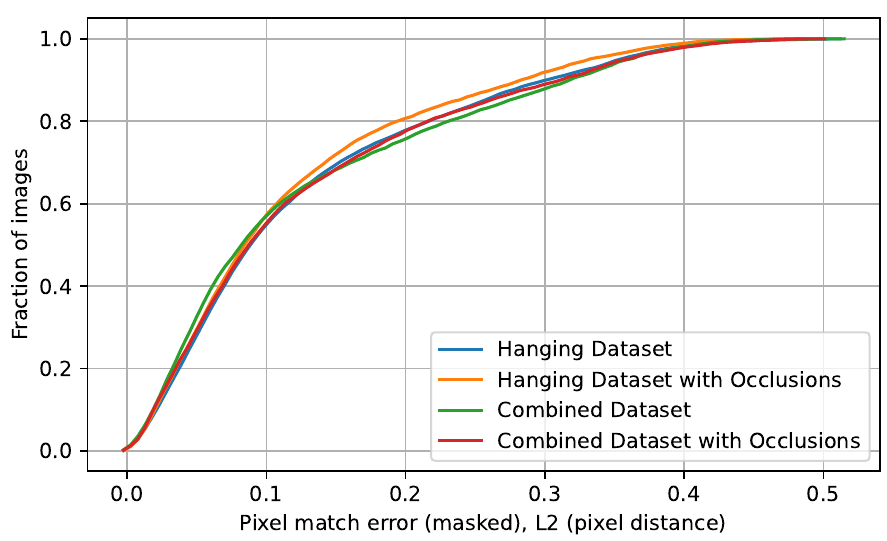}
    \caption{\textbf{Cumulative pixel match error on hanging shirts for networks trained on hanging and combined (hanging and table) datasets with and without occlusions.} The networks all perform similarly in simulation, but we found that on real data, occlusions and the specialized hanging network both performed better.}
    \label{fig:hanging}
\end{figure}

We found that adding artificial occlusions to training images did not seem to impact performance with simulated images (\autoref{fig:hanging}), suggesting that the network was robust to minor occlusions. However, training with occlusions significantly improved performance on real systems, likely due to masking artifacts.

\begin{figure}[h]
    \centering
    \includegraphics[width=0.7\textwidth]{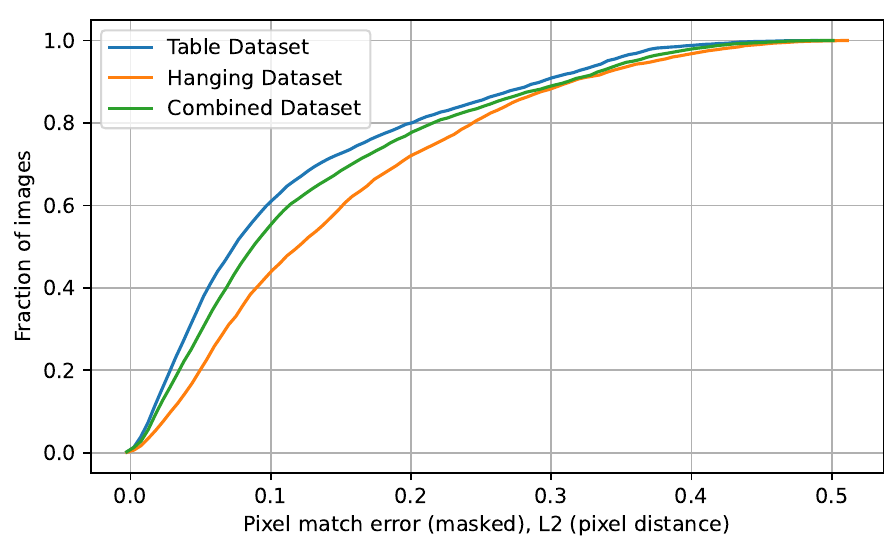}
    \caption{\textbf{Cumulative pixel match error on shirts on a table for networks trained on table, hanging, and combined (hanging and table) datasets.} As hypothesized, the specialty table network performs the best, followed by the network trained with the combined dataset. The hanging network is able to generalize its understanding to shirts on tables, but to a lesser degree of accuracy.}
    \label{fig:table}
\end{figure}

We compare performance of networks trained on exclusively hanging or table scenes to networks trained on a combined dataset (\autoref{fig:hanging}, \ref{fig:table}). The combined network performs marginally worse in both test sets compared to the specialized networks, but does not have significant performance loss. We found that simplifying table configurations during training to be more representative of those used in related works was necessary for improving the combined network's performance. The harder table training set had few distinguishing features, making correspondences more difficult to learn.

\newpage
\subsection{Dense Correspondence Evaluation}
\label{subsec:7.4}
\begin{figure}[t]
    \centering
    \includegraphics[width=\textwidth]{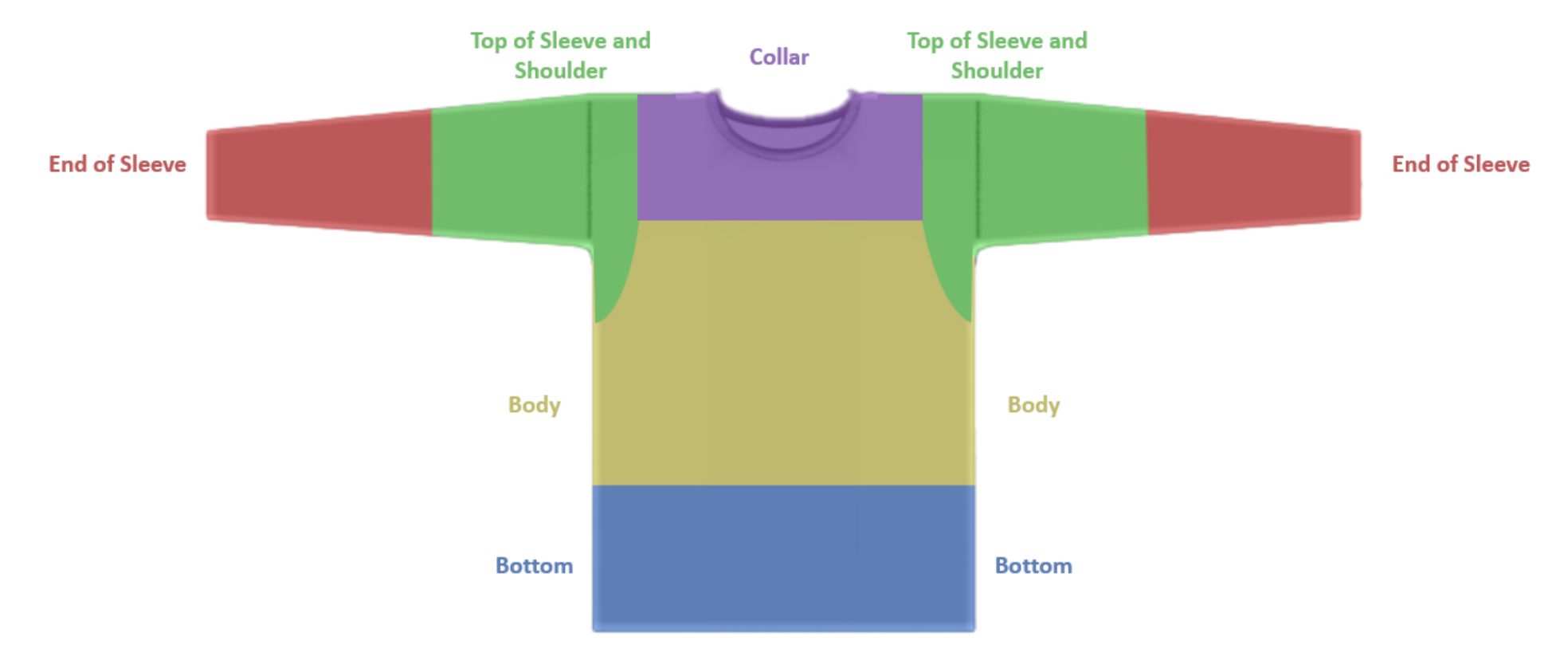}
    \caption{\textbf{Shirt region classification used for real-world evaluation.} During real-world evaluation of the dense correspondence network, a predicted grasp is considered correct if it falls within the same region as the predefined, ground-truth label.}
    \label{fig:shirt_region_classification}
\end{figure}

Our dataset and every single experiment include long-sleeve shirts, varied necklines, and a broad range of materials (including silk blends, stretchy spandex, fuzzy sweaters, etc.).
We evaluate the real-world performance of our dense correspondence network using the color-coded regional classifications defined in \autoref{fig:shirt_region_classification}. In both folding and hanging scenarios, multiple grasp points can lead to the successful execution of a given strategy. Instead of requiring exact pixel-level matches, we divide the shirt into five regions and consider a trial successful if the network's high-confidence grasp prediction falls within the correct region on the physical shirt. 

We conducted ROC studies to help determine optimal correspondence thresholds in simulation (where pixel-level ground truth is available). However, we noted that real-world transfer introduced high
variability. In practice, we found that a confidence threshold of $6 \times 10^{-6}$ reflects a clear inflection point where networks begin to assign high confidence to meaningful regions in real images. Individual pixel confidences peak at approximately $9 \times 10^{-6}$. Low confidence classifications are considered incorrect, but safe. To test in the forward direction (querying on the deformed shirt), we label query points while collecting images. In the inverse direction (querying from the canonical), we query collar, shoulder, sleeve, and bottom points and visualize high confidence matches across all images in the dataset. Points that can be verified or rejected by a human are included in evaluation. Note that not every point is visible in the inverse queries, making low-confidence the ideal option.

We evaluate the accuracy of our dense correspondence network—trained on the combined hanging in-air and table configurations—when picking from the table by determining whether the high-confidence first grasp point the system chooses is within the appropriate region, as defined in \autoref{fig:shirt_region_classification}. We conduct 20 trials to evaluate the network's correspondence prediction success. The configurations of the shirt when picked from the table demonstrate a similar, if not more difficult, deformation as in \cite{Ganapathi2020} and \cite{Unigarment}. Our method shows a comparable success rate to prior works, with the added capability of choosing grasp points from a highly deformed shirt hanging in air. 

\vspace{3mm}
\begin{figure}[H]
    \centering
    \includegraphics[width=\textwidth]{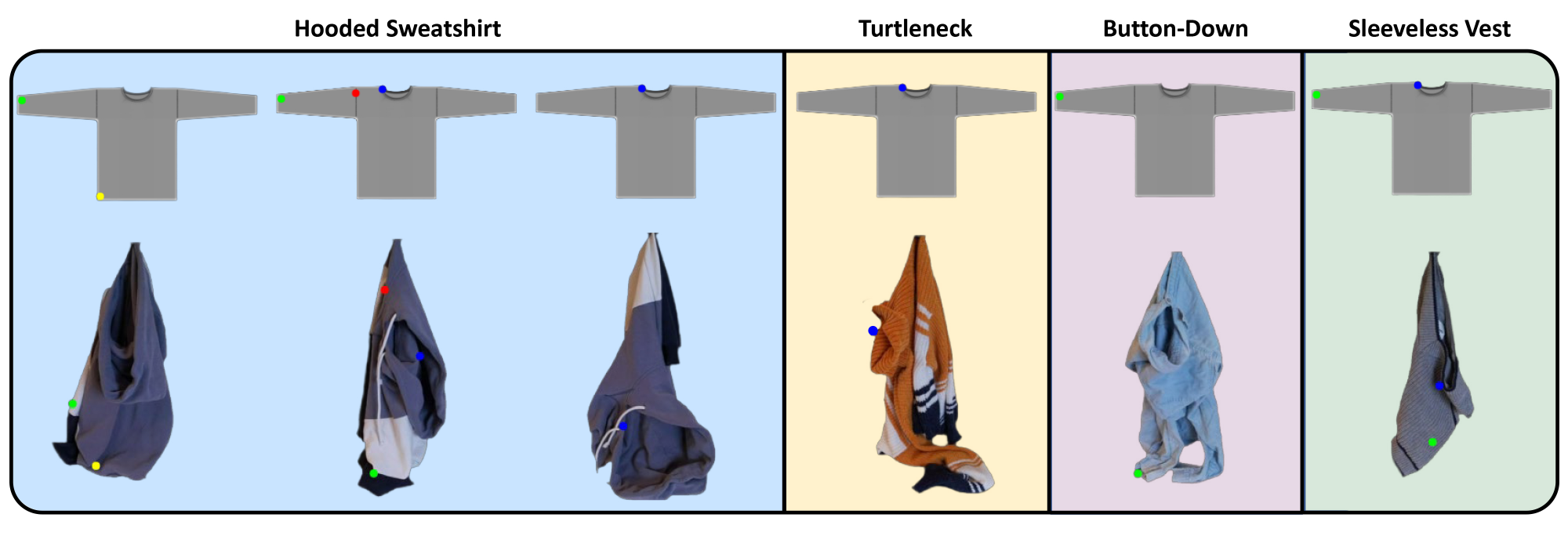}
    \caption{\textbf{Examples of out-of-distribution shirts tested.} We assess the zero-shot out-of-distribution generalization capabilities of our network by testing its predictions in the inverse direction on unseen shirt styles. In general, features such as hoods, turtleneck collars, and buttons not present in the simulated training dataset do not degrade the network's performance, as it is still able to classify shirt features accurately. Some misclassifications do occur with sleeveless shirts, as the network predicts the bottom of the shirt as the end of the sleeve. Overall, the network successfully generalizes to previously unseen shirt styles, demonstrating a visual understanding of the shirt structure.}
    \label{fig:out_of_distribution}
\end{figure}

The dataset simulated in Blender offers much flexibility in rendering a wide range of shirt geometries and details, including variations in body and sleeve length and shirt details. However, features such as hoods, turtleneck collars, buttons, and sleeveless shirts are not simulated. We assess our dense object network's zero-shot generalization capabilities to out-of-distribution shirts in the inverse direction. Notably, previously unseen visual features such as hoods, turtlenecks, and button-up collars do not seem to degrade the network's ability to distinguish the collar regions from the sleeves or bottoms of the shirts. Similarly, color-blocked patterns and buttons do not confuse the network, likely due to the wide range of textures and colors present in the simulated training data. Occasional misclassifications occur with sleeveless shirts and vests, where the network incorrectly predicts the shirt bottom as a sleeve when queried from the canonical shirt. We note, however, this error is also observed in some in-distribution examples. Overall, despite the unseen shirt types, our network demonstrates a general visual understanding of the shirt structure and effectively generalizes to styles beyond those seen in training. See \autoref{fig:out_of_distribution} for examples.

\subsection{Visuotactile Grasp Affordance}
\label{subsec:7.5}
% check if there's anything missing from tactile classifier
% describe affordance criteria in sim in detail. 
% put in images of affordance prediction across three networks

\begin{figure}[H]
    \centering
    \includegraphics[width=\textwidth]{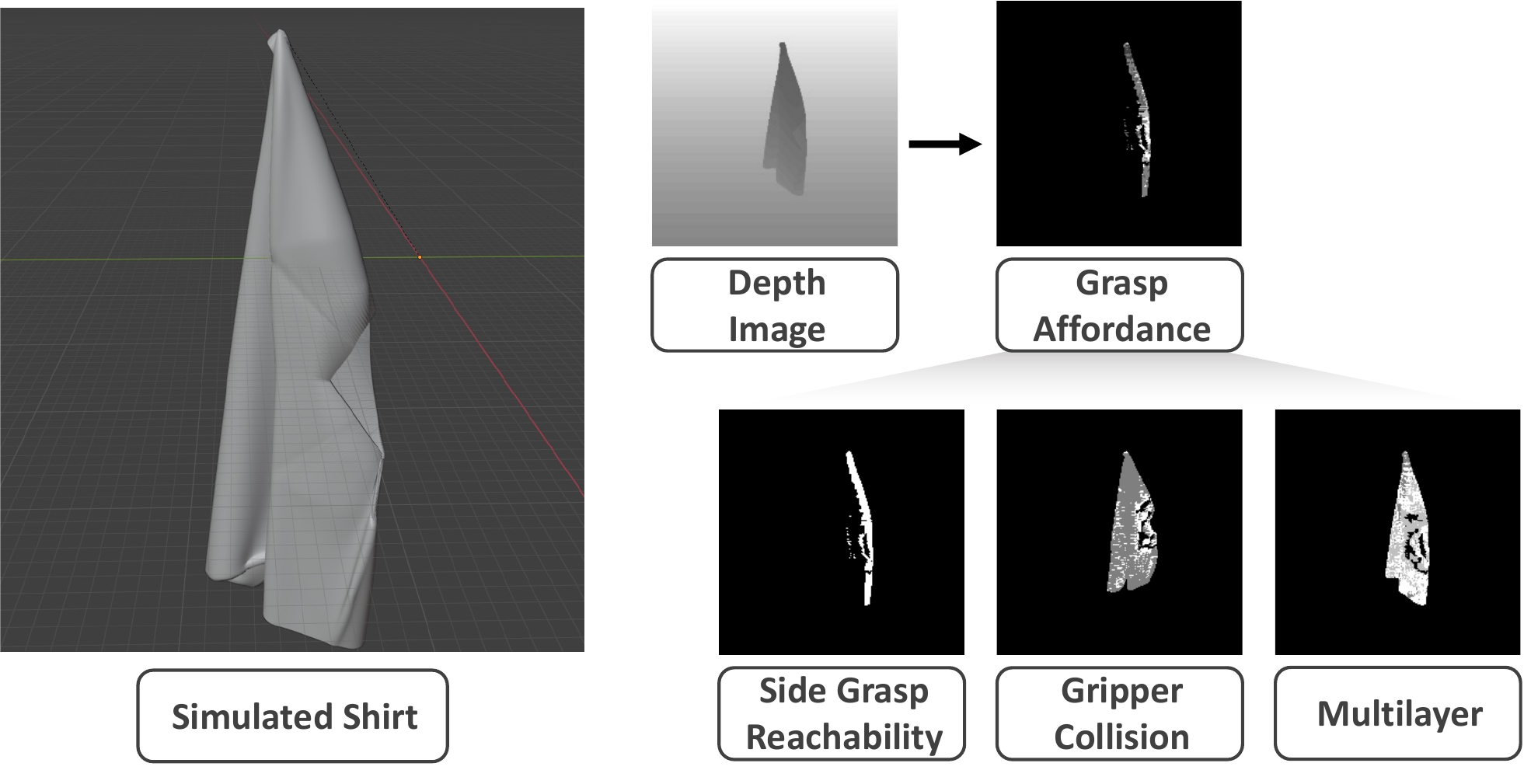}
    \vspace{-5mm}
    \caption{\textbf{Visuotactile grasp affordance training in simulation.} We generate affordance labels for entire images in simulation by evaluating grasp feasibility based on reachability with a side grasp, collision avoidance, and fabric layer count (restricted to two or fewer). We adapt the affordance data generation pipeline introduced in \cite{Visuotactile} to our simulation environment to obtain the affordance labels.}
    \label{fig:training_affordance_in_simulation}
\end{figure}

We compute per-pixel grasp affordance labels in simulation using an adapted version of the method from \cite{Visuotactile}. In our case, the goal is to identify viable side grasps for grasping shirts rather than edge grasps for towels, so we modify the criteria accordingly. Specifically, we remove the edge constraint used in the original formulation and allow up to two fabric layers instead of one. Affordance labels are computed by evaluating whether a candidate grasp point (1) is reachable by the right arm, (2) avoids collision with the cloth during the approach, and (3) results in no more than two layers of fabric between the gripper fingers. \autoref{fig:training_affordance_in_simulation} shows examples of the resulting simulation affordance labels. The network took under 2 hours to train on the simulated network on a Titan X Pascal GPU. 

Collecting 8000 grasps on the robot supervised with our tactile classifier took approximately 14 hours. 
The tactile classifier cannot reliably determine whether the grasped region corresponds to the intended visual target. As a result, non-reachable pixels can yield positive tactile signals due to inadvertently grasping cloth in front of the target. To help address these challenges, we incorporate several training strategies:

\begin{itemize}
    \item \textbf{Neighboring Pixel Loss with Gaussian Weighting:}  
    To enhance robustness to small positional shifts, we include a neighborhood of pixels around the ground-truth grasp point. Each neighboring pixel's loss is weighted by a Gaussian function of its distance from the center:
    {
    \setlength{\abovedisplayskip}{0pt}%
    \setlength{\belowdisplayskip}{0pt}%
    \setlength{\abovedisplayshortskip}{0pt}%
    \setlength{\belowdisplayshortskip}{0pt}%
    \vspace{-1mm}
    \begin{equation}
    \mathcal{L}_{\text{neighbor}} = \frac{1}{|\mathcal{N}|} \sum_{(i,j) \in \mathcal{N}} \exp\left( -\frac{d_{ij}^2}{2\sigma^2} \right) \left( A(i,j) - y_{\text{gt}} \right)^2
    \end{equation}
    }
    where:
    \begin{itemize}
        \item \(\mathcal{N}\) is the set of neighboring pixels within a \(d_{\text{px}} \times d_{\text{px}}\) window centered on the ground-truth grasp point, excluding any pixels outside image bounds.
        \item \(d_{ij}^2 = (i - i_0)^2 + (j - j_0)^2\) is the squared Euclidean distance from the center pixel \((i_0, j_0)\).
        \item \(\sigma\) controls the spread of the Gaussian weighting.
        \item \(|\mathcal{N}|\) is the number of valid neighboring pixels for normalization.
    \end{itemize}

    \vspace{2mm}
    \item \textbf{Spatial Regularization:}  
    Encourages smoothness in the affordance map by penalizing large gradients between adjacent pixels:
    \vspace{1mm}
    {
    \setlength{\abovedisplayskip}{2pt}%
    \setlength{\belowdisplayskip}{2pt}%
    \setlength{\abovedisplayshortskip}{2pt}%
    \setlength{\belowdisplayshortskip}{2pt}%
    \begin{equation}
    \mathcal{L}_{\text{spatial}} = \sum_{i,j} \left| A(i+1, j) - A(i,j) \right| + \left| A(i, j+1) - A(i, j) \right|
    \end{equation}
    }

    \item \textbf{Simulation Regularization:}  
    Ensures consistency between the fine-tuned real-world affordance network and the pretrained simulation network to maintain global structure even with unexplored points:
    \vspace{-1mm}
    {
    \setlength{\abovedisplayskip}{2pt}%
    \setlength{\belowdisplayskip}{2pt}%
    \setlength{\abovedisplayshortskip}{2pt}%
    \setlength{\belowdisplayshortskip}{2pt}%
    \begin{equation}
    \mathcal{L}_{\text{sim}} = \left\| A_{\text{real}} - A_{\text{sim}} \right\|^2
    \end{equation}
    }
    \vspace{1mm}
    \item \textbf{Weight Decay:}  
    Applies L2 regularization to the network weights directly in the optimizer:

    {
    \setlength{\abovedisplayskip}{2pt}%
    \setlength{\belowdisplayskip}{2pt}%
    \setlength{\abovedisplayshortskip}{2pt}%
    \setlength{\belowdisplayshortskip}{2pt}%
    \begin{equation}
    \mathcal{L}_{\text{weight}} = \lambda_{\text{weight}} \left\| \theta \right\|^2
    \end{equation}
    }
\end{itemize}

\begin{figure}[b!]
    \vspace{-5mm}
    \centering
    \includegraphics[width=\textwidth]{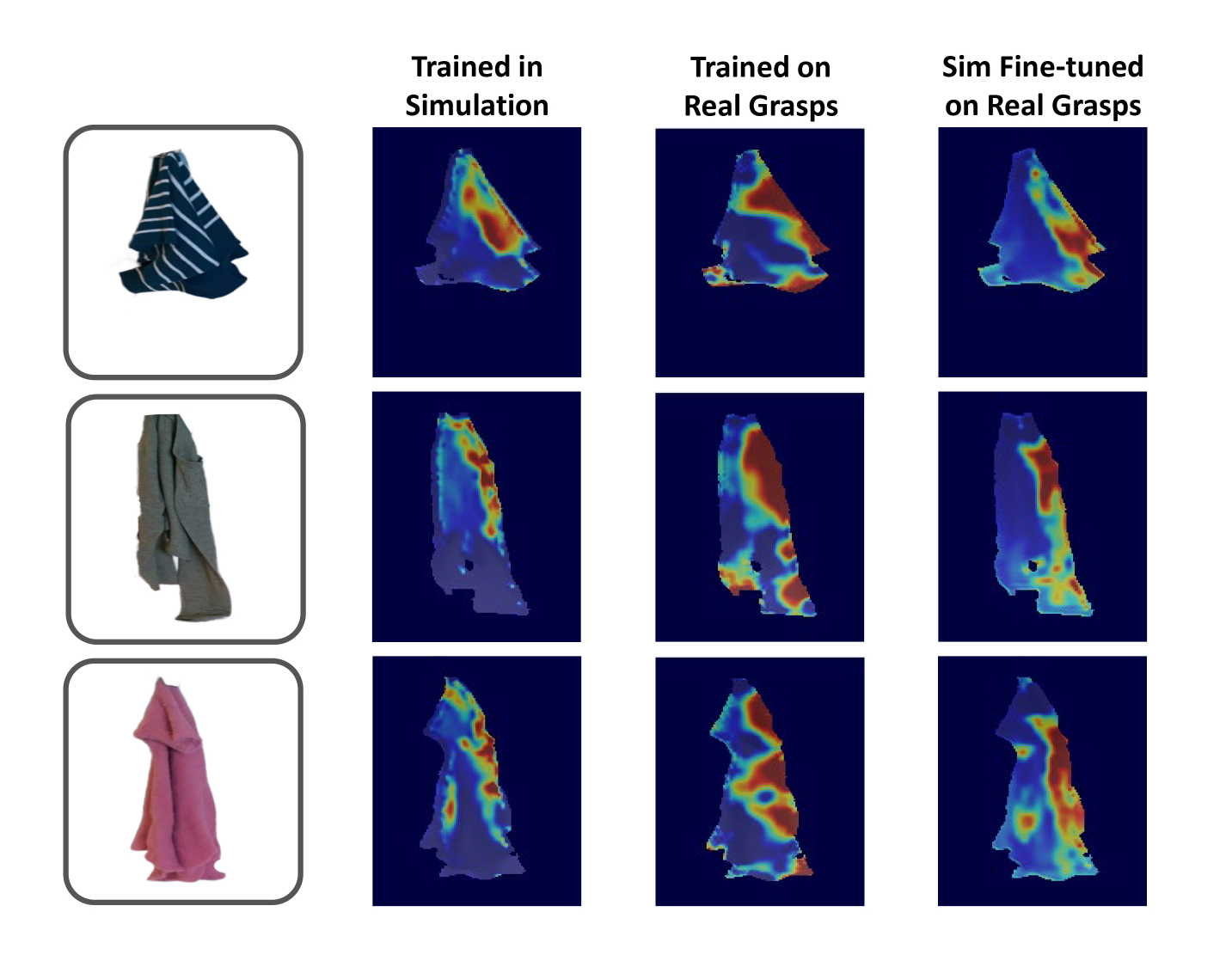}
    \caption{\textbf{Fine-tuned visuotactile grasp affordance compared to baselines.} The model trained in simulation (left, Sim2Real) is overly conservative, often failing to identify viable grasp points—particularly near the bottom of the shirt. In contrast, the model trained only on real robot grasps (middle, Real2Real) is overconfident in unexplored regions and is sensitive to misclassified grasps where the robot contacts fabric inside the shirt, rather than the intended target region, without regularization from the network trained in simulation.}
    \label{fig:shirt_regions}
\end{figure}

\noindent
The total loss is defined as:
\vspace{1mm}
{
\setlength{\abovedisplayskip}{2pt}%
\setlength{\belowdisplayskip}{2pt}%
\setlength{\abovedisplayshortskip}{2pt}%
\setlength{\belowdisplayshortskip}{6pt}%
\begin{equation}
\mathcal{L}_{\text{total}} = \mathcal{L}_{\text{neighbor}} + \lambda_{\text{spatial}} \mathcal{L}_{\text{spatial}} + \lambda_{\text{sim}} \mathcal{L}_{\text{sim}}
\end{equation}
}

\autoref{fig:shirt_regions} compares affordance predictions from networks trained in simulation and on real robot grasps.

\subsection{Human Video Demonstrations}
\label{subsec:7.6}
In order to extract grasp points from human video demonstrations, we trained a custom gesture recognizer based on MediaPipe's GestureRecognizer framework. This network allows us to track transitions between open and grasping hands and tracks the hand skeleton. We identify grasp events as frames in which both hands are in a grasping pose, and extract the first frame of these segments as key frames. The index fingertip of the lower hand is then used as a query point for our dense correspondence model to localize the intended grasp location on a canonical garment image (\autoref{fig:human_demo}). We apply a Segment Anything-based mask \cite{segmentanything} to isolate the garment in the demonstration image.

While the full pipeline enables generalization across different users and environments, its success rate is currently limited. The gesture recognizer can misclassify ambiguous hand poses and the off-the-shelf skeleton tracker occasionally fails to accurately localize the hands. Additionally, the dense correspondence model struggles in frames where the hand occludes the target grasp point. To mitigate occlusion, we select a frame a few steps prior to the grasp, but in many cases, the cloth shifts between these frames, leading to inaccurate grasp localization. This pipeline is outside of the primary focus of our work, but rather a demonstration of the potential for using dense descriptors to interface with unconstrained human video data. With more focused development, these limitations could likely be addressed—for example, by training a more robust, domain-specific gesture recognizer or incorporating occlusion-aware correspondence networks. Despite its current limitations, this approach illustrates how our descriptor representation enables pick point extraction directly from raw demonstrations—a key step toward scaling data collection for garment manipulation.

% \begin{figure}[h!]
%     \centering
%     \includegraphics[width=\textwidth]{figures/human_demo.pdf}
%     \vspace{-1mm}
%     \caption{\textbf{Extracting grasp points from human video demonstrations.} We track hand gestures throughout the video to identify key moments. For each key frame, we use the tracked hand position to define a query point and retrieve the corresponding location on the canonical shirt using our dense correspondence model. This approach enables folding demonstrations to be interpreted as robot-executable instructions via our dense visual representation.}
%     \label{fig:human_demo}
% \end{figure}

%===============================================================================
\end{document}